\documentclass[letterpaper, 10 pt, conference]{ieeeconf}  

\IEEEoverridecommandlockouts                              

\usepackage{times} 
\usepackage{amsmath} 
\usepackage{amssymb}  
\usepackage{graphicx}
\usepackage{booktabs}
\usepackage{subfig}
\usepackage{tabularx}
\usepackage{lipsum}
\usepackage{multirow}
\usepackage{gensymb}
\usepackage{makecell}
\usepackage[pagebackref=true,breaklinks=true,letterpaper=true,colorlinks,bookmarks=false]{hyperref}

\title{\LARGE \bf
Modeling Affect-based Intrinsic Rewards for Exploration and Learning
}

\author{Dean Zadok$^{1,2*}$\thanks{*This paper is the product of work during an internship at Microsoft Research.}, Daniel McDuff$^2$ and Ashish Kapoor$^2$
\thanks{$^{1}$Computer Science Department, Technion, Haifa, Israel
        {\tt\small deanzadok@campus.technion.ac.il}}%
\thanks{$^{2}$Microsoft Research, Redmond, WA, USA
        {\tt\small \{damcduff,akapoor\}@microsoft.com}}%
}

\begin{document}

\newcommand{\etal}{\textit{et al.}}

\maketitle
\thispagestyle{empty}
\pagestyle{empty}

\begin{abstract}
Positive affect has been linked to increased interest, curiosity and satisfaction in human learning. In reinforcement learning, extrinsic rewards are often sparse and difficult to define, intrinsically motivated learning can help address these challenges. We argue that positive affect is an important intrinsic reward that effectively helps drive exploration that is useful in gathering experiences. 
We present a novel approach leveraging a task-independent reward function trained on spontaneous smile behavior that reflects the intrinsic reward of positive affect. To evaluate our approach we trained several downstream computer vision tasks on data collected with our policy and several baseline methods. We show that the policy based on our affective rewards successfully increases the duration of episodes, the area explored and reduces collisions. The impact is the increased speed of learning for several downstream computer vision tasks.
\end{abstract}

\let\thefootnote\relax\footnotetext{Our implementation is on \href{https://github.com/microsoft/affectbased}{https://github.com/microsoft/affectbased}.}

\section{INTRODUCTION}

Reinforcement learning (RL) is most commonly achieved via policy specific rewards that are designed for a predefined task or goal. Such {\em extrinsic} rewards can be sparse and difficult to define and/or only apply to the task at hand. We are interested in exploring the hypothesis that RL frameworks can be designed in a task agnostic fashion and that this will enable us to efficiently learn general representations that are useful in solving several tasks related to perception in robotics. In particular, we consider {\em intrinsic} rewards that are akin to affect mechanisms in humans and encourage efficient and safe explorations. These rewards are task-independent; thus, the experiences they gather are not specific to any particular activity and can be harnessed to build general representations. Furthermore, intrinsically motivated learning can have advantages over extrinsic rewards as they can reduce the sample complexity by producing rewards signals that indicate success or failure before the episode ends \cite{mcduff2018visceral}.

A key question we seek to answer is how to define such an intrinsic policy. We propose a framework that comprises mechanisms motivated by human affect, that mimic affect experienced by real human subjects. The core insight is that learning agents motivated by drives such as delight, fear, curiosity, hunger, etc. can garner rich experiences that are useful in solving multiple types of tasks. For instance, in reinforcement learning contexts there is a need for an agent to adequately explore its environment~\cite{frank2014curiosity}. This can be performed by randomly selecting actions, or employing a more intelligent strategy, directed exploration, that incentives exploration of unexplored regions~\cite{chen2018learning}. Curiosity is often defined by using the prediction error as the reward signal~\cite{pathak18largescale,pathakICMl17curiosity}. As such, the uncertainty, or mistakes, made by the system are assumed to represent what the system should want to learn more about. However, this is a simplistic view as it fails to take into account that new stimuli are not always very informative or useful~\cite{savinov2018episodic}. Savinov \etal uses the analogy of becoming glued to a TV channel surfing when there is the rest of the world outside the window. Their work proposed a new novelty bonus that features episodic memory. McDuff and Kapoor~\cite{mcduff2018visceral} took another approach focusing on safe exploration, proposing intrinsic rewards mimicking responses to that of a human's sympathetic nervous system (SNS) to avoid catastrophic mistakes during the learning phase, but not necessarily promoting exploration.

\begin{figure*}[t]
\centering
\includegraphics[width = 0.97\textwidth]{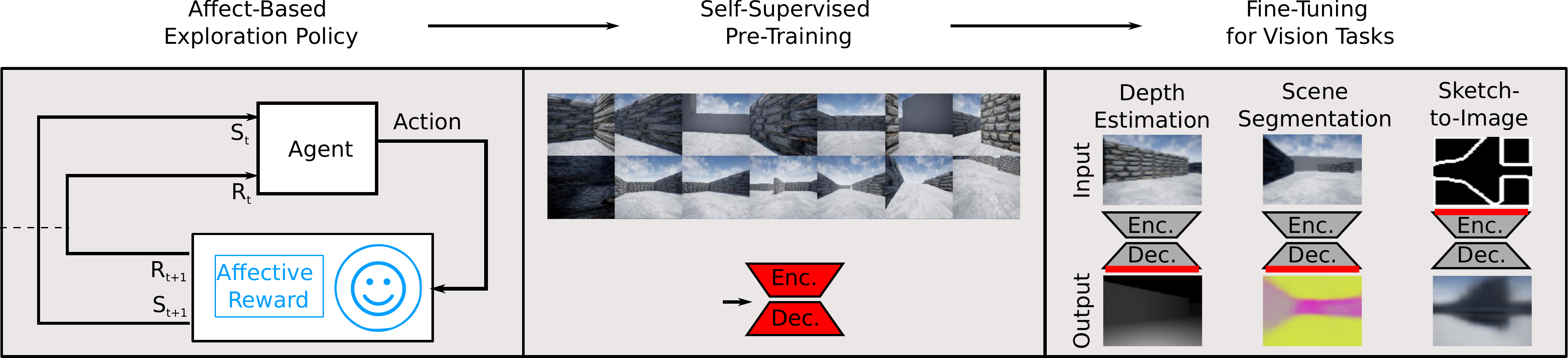} 
\caption{We present a novel approach leveraging a reward that mimics positive affect-based intrinsic rewards to motivate exploration. We use this policy to collect data for self-supervised pre-training and then use the learned representations for multiple downstream computer vision tasks. The red regions highlight the parts of the architecture trained at each stage.}
\label{figure:summary}
\vspace{-0.2cm}
\end{figure*}

In this paper, we specifically focus on the role of positive emotions and study how such intrinsic motivations can enable learning agents to explore efficiently and learn useful representations. 
 In research on education, positive affect has been shown to be related to increased interest, involvement and arousal in learning contexts~\cite{masters1979affective}. Kort \etal's~\cite{kort2001affective} model of emotions in learning posits that the states of curiosity and satisfaction are associated with positive affect in the context of a constructive learning experience. Human physiology is informative about underlying affective states. Smile behavior~\cite{jaques2018learning} and physiological signals~\cite{mcduff2018visceral} have been effectively used as feedback in learning systems but not in the context of intrinsic motivation or curiosity. We leverage facial expressions as an unobtrusive measure of expressed positive affect. The key challenges here entail both designing a system that can first model the intrinsic reward appropriately, and then building a learning framework that can efficiently use the data to solve multiple downstream tasks related to perception in robotics.

The core contributions of this paper (summarized in Fig.~\ref{figure:summary}) are to (1) present a novel learning framework in which reward mechanisms motivated by positive affect-mechanisms in humans are used to carry out explorations while being agnostic to any specific tasks, (2) show how the data collected in such an {\em unsupervised} manner can be used to build general representations useful for solving downstream tasks with minimal task-specific fine-tuning, and (3) report the results on experiments showing that the framework improves exploration as well as enabling efficient learning for solving multiple data-driven tasks. In summary, we argue that such an intrinsically motivated learning framework inspired by affective mechanisms can be effective in increasing the coverage during exploration, decreasing the number catastrophic failures and that the garnered experiences can help us learn general representations for solving tasks including depth estimation, scene segmentation, and sketch-to-image translation.

\section{Related Work}

Our work is inspired by intrinsically motivated learning~\cite{chentanez2005intrinsically,haber2018learning,zheng2018learning}. One key property of intrinsic rewards is that they are non-sparse~\cite{savinov2018episodic}, this helps aid learning even if the signal is weak. Much of the work in this domain uses a combination of intrinsic and extrinsic rewards in learning.  Curiosity is one example of an intrinsic reward that grants a bonus when an agent discovers something new and is vital for discovering successful behavioral strategies. For example,~\cite{pathak18largescale,pathakICMl17curiosity} models curiosity via the prediction error as a surrogate and shows that such intrinsic reward mechanism performed similarly to hand-designed extrinsic rewards in many environments. Similarly, Savinov \etal~\cite{savinov2018episodic} defined a different curiosity metric based on how many steps it takes to reach the current observation from those in memory, thus capturing environment dynamics. Their motivation was that previous approaches were too simplistic in assuming that all changes in the environment should be considered equal. 
McDuff and Kapoor~\cite{mcduff2018visceral} provided an example of how an intrinsic reward mechanism could motivate safer learning, utilizing human physiological responses to shorten the training time and avoid critical states. Our work is inspired by the prior art, but with the key distinction that we specifically aim to build intrinsic reward mechanisms that are visceral and trained on signals correlated with human affective responses.

Imitation learning (IL) is a popular method for deriving policies. In IL, the model is trained on previously generated examples to imitate the recorded behavior. It has been successfully implemented using data collected from many domains~\cite{billard2001learning, blukis2018following,bojarski2016end,cardamone2009learning,ross2011reduction, ross2013learning}. 
Simulated environments have been successfully used for training and evaluating IL systems~\cite{codevilla2018end,zadok2019explorations}. We use IL as a baseline in our work and perform experiments to show how a combination of IL and positive affect-based rewards can lead to greater and safer exploration.

One of our goals is to explore whether our intrinsically motivated policy can help us learn general representations. Fortunately, the rise of unsupervised generative models, such as generative adversarial networks (GANs)~\cite{goodfellow2014generative} and variational auto-encoders (VAE)~\cite{kingma2013auto}, has led to progress across many interesting challenges in computer vision that involve image-to-image translation~\cite{isola2017image,zhu2017unpaired}. In our work, we use three tasks in our data evaluation process: scene segmentation~\cite{DEEPLAB}, depth estimation~\cite{godard2017unsupervised}, and sketch-to-image translation~\cite{chen2018sketchygan}. The first two tasks are common in driving scenarios and augmented reality~\cite{pan2017virtual,swan2007egocentric,wang2019monocular}, while the third is known for helping people render visual content~\cite{eitz2012humans} or synthesize imaginary images~\cite{chen2009sketch2photo}. In this paper, we show that by pre-training a VAE in a self-supervised way first using our exploration policy we can obtain better results on all three tasks.

\section{Our Framework}
\label{sec:ourframework}

Fig.~\ref{figure:summary} describes the overall framework. The core idea behind the proposed methodology is that we model intrinsic motivations in the agents that lead to extensive exploration. Consequently, an agent on its own is able to gather experiences and data in a wide variety of conditions. Note that unlike traditional machine intelligence approaches, the agent is not fixated on a given task - all it is encouraged to do is explore as extensively as possible without getting into perilous situations. The rich data that is being gathered then needs to be harnessed into building representations that will eventually be useful in solving many perceptions tasks. Thus, the framework consists of three core components: (1) A positive-affect based exploration policy, (2) a self-supervised representation learning component and (3) mechanisms that utilize these representations {\em efficiently} to solve various tasks.

\subsection{Affect-Based Exploration Policy}
\label{ssec:affectbasedpolicy}

The approach here is to create a model that encourages the agent to explore the environment. We want a reward mechanism that positively reinforces behaviors that mimics a human's affective responses and lead to discovery and joy. 

\noindent\textbf{Positive Affect-based Reward Mechanism:} 
In our work, we use a Convolutional Neural Network (CNN) to model the affective responses of a human, as if we were in the same scenario as the agent. We train the CNN model to predict human smile responses as the exploration evolves. Based on the fact that positive affect plays a central role in curiosity and learning~\cite{kort2001affective}, we chose to measure smiles as an approximate measure of positive affect. Smiles are consistently linked with positive emotional valence~\cite{brown1980relationships,kassam2010assessment} and have a long history of study~\cite{lafrance2011lip} using electromyography~\cite{brown1980relationships} and automated facial coding~\cite{martinez2017automatic}. We must emphasize that in this work we were not attempting to model the psychological processes that cause people to smile explicitly. We are only using smiles as an outward indicator of situations that are correlated with positive affect as people explore new environments. In particular, the NN was trained to infer the reward $h(x_i, a)$ directly given that an action $a$ was taken when at state $x_i$.
We defer the details of the NN architecture, the process of data collection and the training procedure to Sections~\ref{sec:exp} and~\ref{sec:appendix}.

\noindent\textbf{Choosing Actions with Intrinsic Rewards: }
Given the reward mechanism, we can use any off-the-shelf sequential decision-making framework, such as RL~\cite{mcduff2018visceral}, to learn a policy. It is also feasible to modify an existing policy that is trained to explore or collect data. While the former approach is a desirable one theoretically, it requires a very large number of training episodes to return a useful policy \cite{kaelbling1996reinforcement,kober2013reinforcement}. We focus on the later, where we assume there exists a function $f(x_i)$ which can predict a vector of actions probabilities $p(a)$ when the agent observes state $x_i$. Formally, given an observation $x_i$ and a model $f$, the next action, $a_{i+1}$, is gets selected as: $a_{i+1} = \arg \max_{a} f(x_i)$.
Such a function can be trained on human demonstrations while they explore the environment.

We then use the positive affect model to change the action selection such that it biases the actions that promise to provide better intrinsic rewards. Intuitively, instead of simply using the output of the pre-trained policy $f$ to decide on the next action, we consider the impact of the intrinsic motivation for every possible action consideration. Formally, given the positive affect model $h$, a pre-trained exploration policy $f$, observation $x_i$, the next action, $a_{i+1}$, being selected becomes:
\begin{equation}
a_{i+1} = \arg \max_{a} \{ f(x_i) + \gamma h(x_i, a) \}
\end{equation}
The above equation adds a weighted intrinsic motivation component to the action probabilities from the original model $f$. The weighting parameter $\gamma$ defines the trade-off between the original policy and the effect of the affect-based reward.

\begin{figure*}[t]
\centering
\includegraphics[width = 0.97\textwidth]{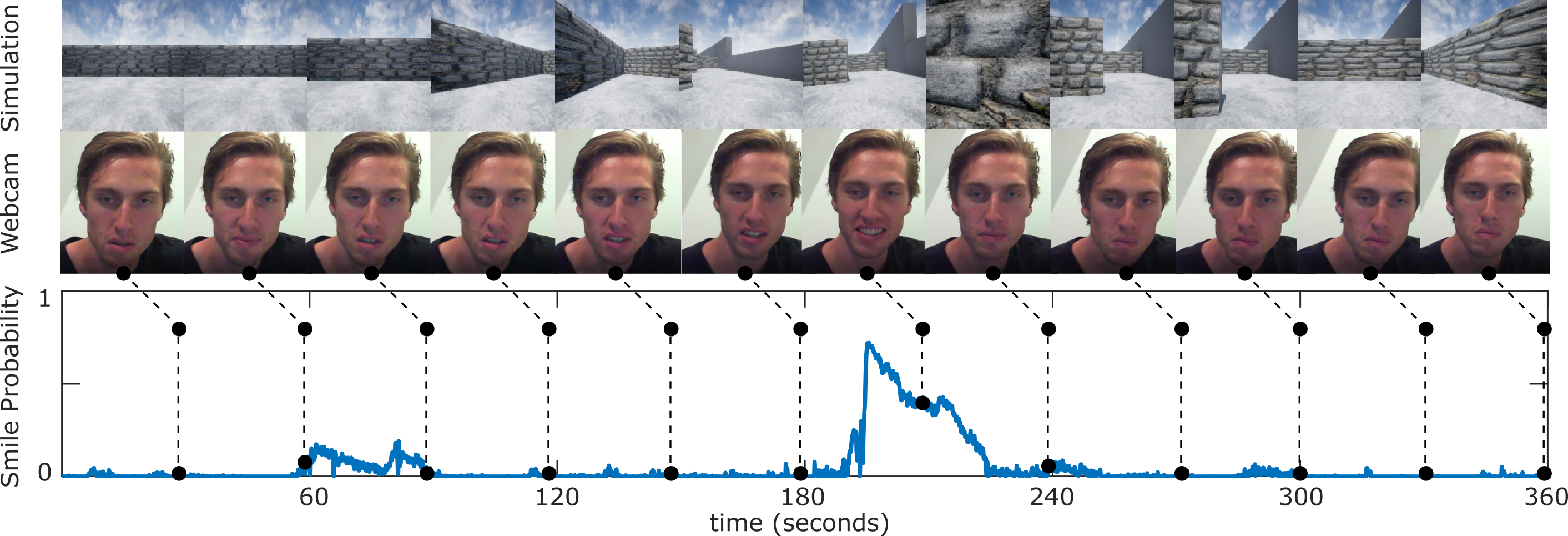} 
\caption{An example of the smile response for a six-minute (360s) period during one of the driving sessions. Frames from the environment and from the webcam video are shown as a reference.}
\label{figure:userstudydata}
\end{figure*}

\subsection{Self-Supervised Learning}
\label{ssec:selfsupervisedlearning}
Given the exploration policy, the agent has the ability to explore and collect rich data. The next component aims to use this data to build rich representations that eventually could be used for various visual recognition and understanding tasks.

The challenge here is that since the collected data was task agnostic there are no clear labels that could be used for supervised learning. We consequently, use the task of jointly learning an autoencoder and decoder, through a low-dimensional latent representation. Formally, we use a variational autoencoder (VAE) to build such representations. For example, a VAE can be trained to restore just the input image, with the loss constructed as the combination of negative log-likelihood and KL divergence, as follows:
\begin{eqnarray}
\begin{split}
L(\theta,\phi)=-\mathbb{E}_{z \sim q_{\theta}(z|x)}[\log p_{\phi}(x|z)] + \\
\mathbb{K}\mathbb{L}(q_{\theta}(z|x)||p_{\phi}(z))
\end{split}
\end{eqnarray}
Where the encoder is denoted by $q_{\theta}$, the decoder is denoted by $p_{\phi}$ and $z$ denotes the low-dimensional projection of the input $x$. The key intuition here is that if the VAE can successfully encode and decode frames then implicitly it is considering aspects such as depth, segmentation, and textures that are critical to making successful predictions. Thus, it should be possible to tweak and fine-tune these VAE networks to solve a host of visual recognition and perception tasks with minimal effort.

\subsection{Fine-tuning for Vision Tasks}
\label{ssec:finetune}
Given the VAE representation, our goal now is to reuse the learned weights to solve standard machine perception tasks. Formally, given some labeled data corresponding to a visual task, similar to supervised learning, we optimize the negative log-likelihood:
\begin{equation}
L(\phi)=-\mathbb{E}_{z \sim q_{\theta}(z|x)}[\log p_{\phi}(x|z)]
\end{equation}
Note that the goal is to minimally modify the network. In our experiments, we show how we can solve depth map estimation and scene segmentation by only tweaking the weights for the first or last few layers just before the decoder output. 
We also show how we can use those weights for sketch-to-image translation, even with a small amount of annotated samples.

\section{Experiments}
\label{sec:exp}

We conducted experiments to analyze (1) the potential advantages of affect-based exploration policy over other heuristics, (2) the ability to learn general representations and (3) if it was possible to solve diverse computer vision tasks by building upon and minimally modifying the model.

\begin{figure*}[t]
\centering
\setlength\tabcolsep{1pt}
\begin{tabular}{cccccc}
Map & Random & Straight & IL & IL +~\cite{mcduff2018visceral} & Affect-based \\ [-1.4ex]
\subfloat{\includegraphics[width = 0.14\textwidth]{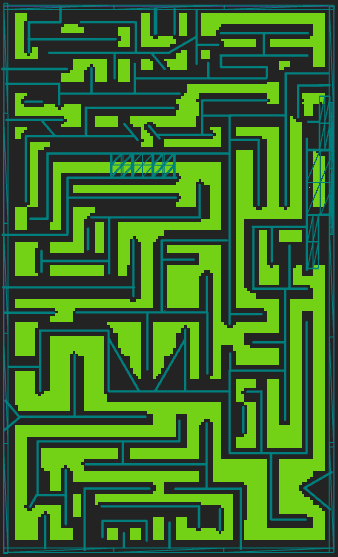}} & 
\subfloat{\includegraphics[width = 0.14\textwidth]{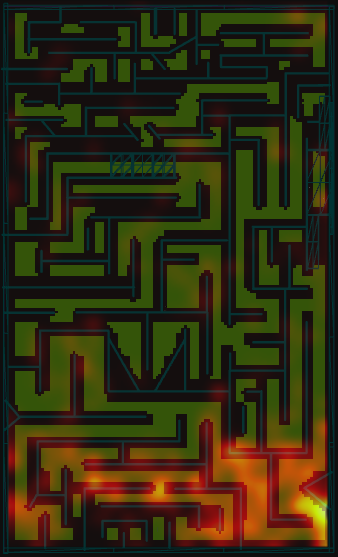}} & 
\subfloat{\includegraphics[width = 0.14\textwidth]{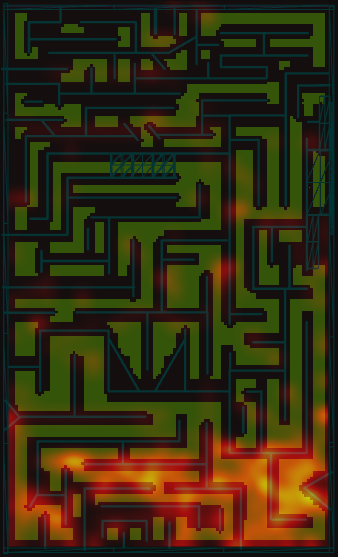}} & 
\subfloat{\includegraphics[width = 0.14\textwidth]{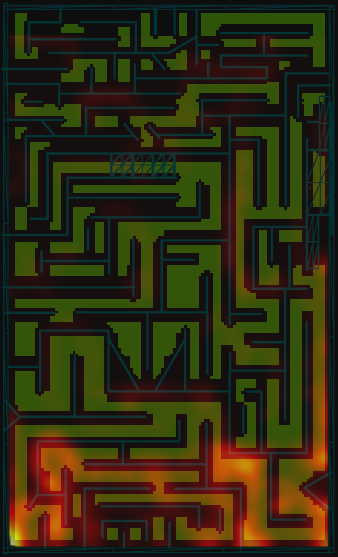}} & 
\subfloat{\includegraphics[width = 0.14\textwidth]{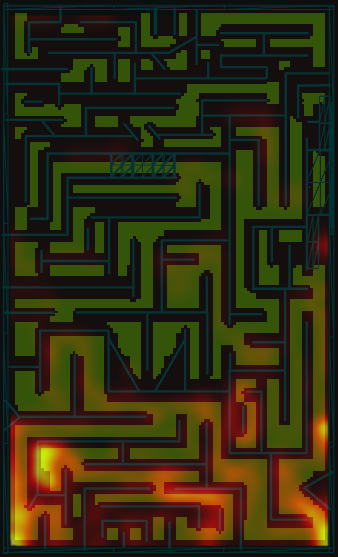}} & 
\subfloat{\includegraphics[width = 0.14\textwidth]{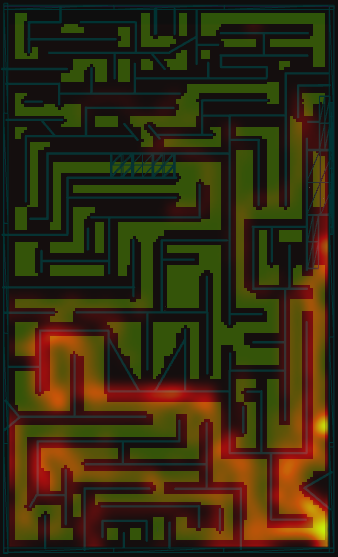}} 
\end{tabular}
\caption{Visualization of the experiment from Table~\ref{tab:cov} using heat maps. From this visualization we can observe that the better the policy, the longer the paths that are recorded during the trials.}
\label{fig:heatmap}
\end{figure*}

We used a high fidelity simulation environment~\cite{shah2018airsim} for autonomous systems, contains a customized 3D maze (dimensions: $2490$ meters by $1500$ meters), a top-down view can be seen in Fig.~\ref{fig:heatmap}. The maze is composed of walls and ramps, frames from the environment can be seen in Fig.~\ref{figure:userstudydata}. The agent we used was a vehicle capable of maneuvering comfortably within the maze. To generate random starting points that allowed us to deploy the agent into the maze, we constructed a navigable area, according to the vehicle dimensions and surroundings (the green region in Fig.~\ref{fig:heatmap}).

\subsection{Data and Model Training}

\noindent\textbf{Positive Affect-based Reward:}
We collected data from five subjects (four males, one female; ages: $23$ - $40$ years) exploring in the simulated environment. All participants were qualified drivers with multiple years of driving experience. Simultaneously we collected synchronized videos of their face. The participants drove for an average of $11$ minutes each, providing a total of approximately $64000$ frames. The protocol was approved by our institutional review board. The participants were told to explore the environment but were given no additional instruction about other objectives. 
We used a well-validated and open-source algorithm to calculate the smile response of the drivers from the webcam videos~\cite{baltrusaitis2018openface}. When evaluated on a large set of videos of naturalistic facial expressions (very similar to ours) the smile detection had a $0.85$ correlation with human coding. An example of a smile response from one subject can be seen in Fig.~\ref{figure:userstudydata}. Using these data we trained our affect-based intrinsic motivation model. The image frames from the camera sensor in the environment served as input to the network and the smile probability in the corresponding webcam frame served as output. The input frames were downsampled to $84 \times 84$ pixels and normalized to be in the range $[0,1]$.
The model architecture is described in Section~\ref{sec:appendix}.

\begin{table*}[t]
\centering
\begin{tabular}{clcccc} 
\toprule
\# & Method & Duration ($s$) & Coverage ($m^2$) & Coverage/sec ($m^2/s$) & Collisions \\ 
\midrule
(1) & Random & 7.57 & 107.79 & \textbf{14.23} & 230 \\
(2) & Straight & 8.32 & 115.33 & 13.86 & 206 \\
(3) & IL & 52.87 & 727.46 & 13.75 & 38 \\
(4) & IL + \cite{mcduff2018visceral} & \textbf{87.63} & 952.82 & 10.87 & \textbf{23} \\
(5) & Affect-based & 79.76 & \textbf{1059.29} & 13.28 & 27 \\
\toprule 
\end{tabular}
\caption{Evaluation of the driving policies. Given a random starting point, duration is the average time the car drove before a collision and coverage is the average area the car covered.}
\label{tab:cov}
\end{table*}

\noindent\textbf{Exploration Policy:}
We first train the base policy $f$ by imitation learning, where data was recorded by a human driver that was trained to drive in the simulation using the vehicle. The data set has $50000$ images, which were normalized to $[0, 1]$ and corresponding human actions.
The model is a CNN trained to classify the desired steering angle. The input space of the model contains four consecutive images, down-sampled to $84 \times 84$, similar to many DQN applications~\cite{mnih2013playing}. The action space is discrete and composed of five possible different steering angles: $40$, $20$, $0$, $-20$ and $-40$\degree. 
The model architecture is described in Section~\ref{sec:appendix}.
To increase the variations in the collected data and cope better with sharp turns, we shifted sampled frames and post-processed the steering angle accordingly~\cite{zadok2019explorations}.
The final exploration policy embeds this $f$ as described in Section~\ref{ssec:affectbasedpolicy}, and considers the affective-rewards. Specifically, the reward mechanism was computed for each one of the steering angles, so the positive intrinsic values represent the values inferred from looking directly towards the respective driving directions. We set $\gamma=6$ which was determined via cross-validation, as described in Section~\ref{sec:appendix}.

\noindent\textbf{Representation Learning:}
As the vehicle explores the environment the data it sees is used to train the VAE. Each episode was initiated by placing the vehicle at a random starting point and letting it drive until the collision. Here we use the task of frame restoration to train the VAE model to restore down-sampled $64 \times 64$ images.
The model architecture is described in Section~\ref{sec:appendix}.
For evaluating performance on depth map estimation and scene segmentation, we collected $2000$ images with ground truth, captured by placing simulated cameras randomly in the environment. For sketch-to-image translation, we used the same method except that the sketches were computed by finding the image contours.

\begin{figure*}[t]
\centering
\includegraphics[width = 0.9\textwidth]{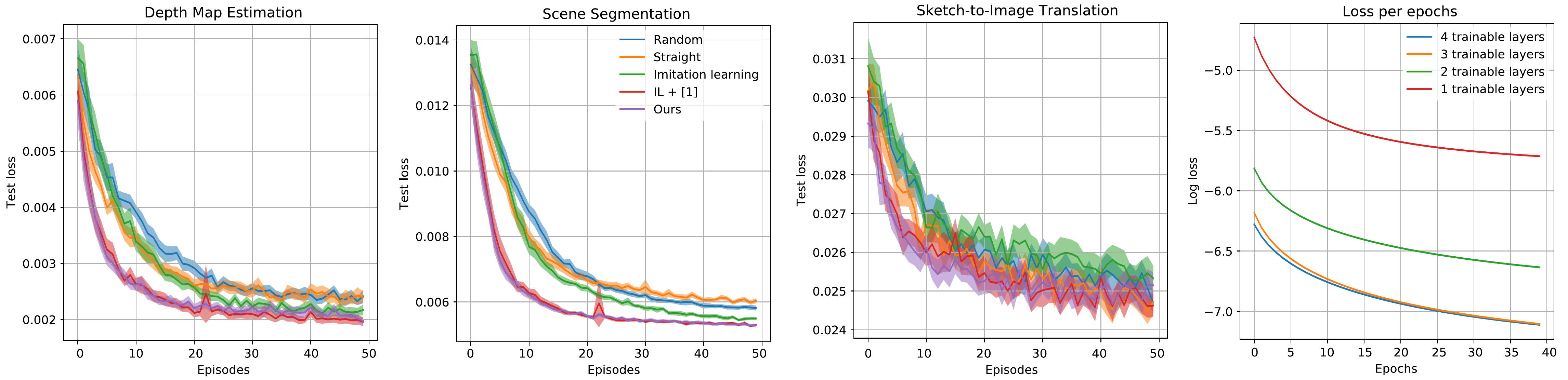} 
\caption{(Left) Test loss as a function of a number of episodes for depth map estimation, scene segmentation, and sketch-to-image translation. Results are averaged over 30 trials. The error bars reflect the standard error. (Right) Test loss as we vary the number of layers tuned for scene segmentation. Fine-tuning just a couple of layers is enough for this task.}
\label{figure:phase2}
\vspace{-.2cm}
\end{figure*}

\begin{figure*}[t]
\begin{center}
\setlength\tabcolsep{0pt}

\begin{tabular}{>{\centering\arraybackslash}ccccccccccccc}
 & \text{ }\text{ }\text{ } & \multicolumn{3}{c}{Depth Map Estimation} & \text{ } & \multicolumn{3}{c}{Scene Segmentation} & \text{ } & \multicolumn{3}{c}{Sketch-to-Image Translation} \\ [-1.5ex]
\rotatebox[origin=l]{90}{\text{ }\text{ }\text{ }Input} & & 
\subfloat{\includegraphics[width = 0.10\textwidth]{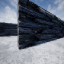}} & 
\subfloat{\includegraphics[width = 0.10\textwidth]{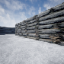}} & 
\subfloat{\includegraphics[width = 0.10\textwidth]{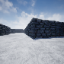}} & & 
\subfloat{\includegraphics[width = 0.10\textwidth]{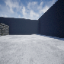}} & 
\subfloat{\includegraphics[width = 0.10\textwidth]{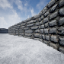}} & 
\subfloat{\includegraphics[width = 0.10\textwidth]{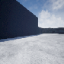}} & & 
\subfloat{\includegraphics[width = 0.10\textwidth]{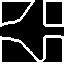}} & 
\subfloat{\includegraphics[width = 0.10\textwidth]{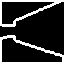}} & 
\subfloat{\includegraphics[width = 0.10\textwidth]{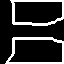}} \\ [-2.5ex]
\rotatebox[origin=l]{90}{\text{ }\text{ }\text{ }\text{ }\text{ }GT} & & 
\subfloat{\includegraphics[width = 0.10\textwidth]{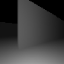}} & 
\subfloat{\includegraphics[width = 0.10\textwidth]{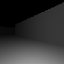}} & 
\subfloat{\includegraphics[width = 0.10\textwidth]{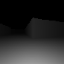}} & & 
\subfloat{\includegraphics[width = 0.10\textwidth]{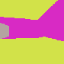}} & 
\subfloat{\includegraphics[width = 0.10\textwidth]{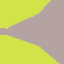}} & 
\subfloat{\includegraphics[width = 0.10\textwidth]{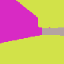}} & & 
\subfloat{\includegraphics[width = 0.10\textwidth]{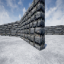}} & 
\subfloat{\includegraphics[width = 0.10\textwidth]{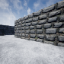}} & 
\subfloat{\includegraphics[width = 0.10\textwidth]{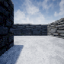}} \\ [-3.1ex]
\rotatebox[origin=l]{90}{\text{ }\text{ }\text{ }\text{ }\text{ }IL} & & 
\subfloat{\includegraphics[width = 0.10\textwidth]{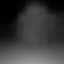}} & 
\subfloat{\includegraphics[width = 0.10\textwidth]{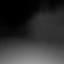}} & 
\subfloat{\includegraphics[width = 0.10\textwidth]{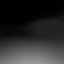}} & & 
\subfloat{\includegraphics[width = 0.10\textwidth]{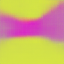}} & 
\subfloat{\includegraphics[width = 0.10\textwidth]{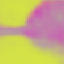}} & 
\subfloat{\includegraphics[width = 0.10\textwidth]{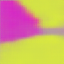}} & & 
\subfloat{\includegraphics[width = 0.10\textwidth]{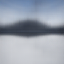}} & 
\subfloat{\includegraphics[width = 0.10\textwidth]{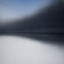}} & 
\subfloat{\includegraphics[width = 0.10\textwidth]{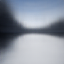}} \\ [-3.1ex]
\rotatebox[origin=l]{90}{\text{ }\text{ }Affect} & & 
\subfloat{\includegraphics[width = 0.10\textwidth]{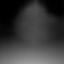}} &
\subfloat{\includegraphics[width = 0.10\textwidth]{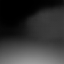}} &
\subfloat{\includegraphics[width = 0.10\textwidth]{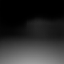}} & & 
\subfloat{\includegraphics[width = 0.10\textwidth]{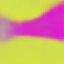}} & 
\subfloat{\includegraphics[width = 0.10\textwidth]{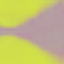}} & 
\subfloat{\includegraphics[width = 0.10\textwidth]{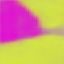}} & & 
\subfloat{\includegraphics[width = 0.10\textwidth]{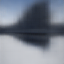}} & 
\subfloat{\includegraphics[width = 0.10\textwidth]{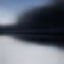}} & 
\subfloat{\includegraphics[width = 0.10\textwidth]{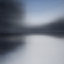}} \\ [-1.5ex]
\end{tabular}
\end{center}
\caption{Samples generated using VAEs trained on each of the three tasks. Notice that there are fewer distortions in the depth map estimations, better classification of structures in the scene segmentation, and better generation of the images from sketches when using our proposed policy.}
\label{fig:samples}
\vspace{-0.5cm}
\end{figure*}

\subsection{Results}
\label{ssec:results}

\noindent{\bf How good is affect-driven exploration?}
We compare the proposed method to four additional methods - random, straight, IL and IL +~\cite{mcduff2018visceral}. For the random policy, we simply draw a random action for each timestamp according to a uniform distribution, regardless of the input space. For the straight policy, the model drives straight, without changing course. The IL policy is simply the base policy $f$, but without the intrinsically motivated reward component. For~\cite{mcduff2018visceral}, to compare their method with ours, we combined their intrinsic reward function with the base policy, the same as we did in ours. When implementing such baselines using the RL approach, we achieved significantly lower results for driving duration.

For this experiment, we select a starting point randomly from the navigable area and let the policy drive until a collision is detected. Then, we reset the vehicle to a random starting position. We continue this process for $2000$ seconds, with the vehicle driving at a speed of $2.5 m/s$. We then consider the mean values of the duration and of the total area covered during exploration per episode. Longer episodes reflect that the policy is able to reason about free spaces and consider that with the vehicle dynamics, whereas higher coverage suggests that the policy indeed encourages novel experiences and that the vehicle is not simply going in circles (or stationary). Coverage is defined as a union of circles, with a $3$-meter radius centered around the car.

Fig.~\ref{fig:heatmap} shows the map of the environment and how different policies explore the space. The heat signature (yellow is more), indicates the amount of time a vehicle spends at that location. We observe that our policy driven by positive affect-based reward is able to go further and cover significantly bigger space in the allocated $2000$ seconds. We present the numerical results in Table~\ref{tab:cov}, which shows that despite being~\cite{mcduff2018visceral} better at holding onto the track, it is less good in exploring the field and cover less area per seconds in average.

\noindent{\bf How well can we solve other tasks?}
In this experiment, we explore how well the VAE trained on the task-agnostic data collected via the exploration (as described in Section~\ref{ssec:selfsupervisedlearning}) can help solve depth-map estimation, scene segmentation, and sketch-to-image translation. First, we explore how much do we really need to perturb the original VAE model to get reasonable performance on these tasks, with only retraining the top few layers of the decoder. In Fig.~\ref{figure:phase2} (right) we plot the log loss, where loss is the task-specific reconstruction L2 loss, as the training evolves over several epochs. The figure shows curves when we choose different numbers of layers to retrain. What we find is that the biggest gains in performance happen with just the top-2 layers. 
Note that since the loss is in log-scale the difference in performance between training the top-2 layers and retraining a higher number of layers is small. Note that for the sketch-to-image translation it was necessary to retrain the encoder as the input sketches differ from images.

Next, we also study the effect of exploration policies on the three tasks. For these experiments, as the vehicle explores the environment, we in-parallel train the VAE model on the data gathered so far and measure the performance on the three-tasks. 
We report our results by averaging over $30$ trials. Fig.~\ref{figure:phase2} (left) shows the mean test L2 loss as the number of episodes evolves for various exploration policies. 
While for most of the policies we observe that the error goes down with the number of episodes, for affect-based policy achieves lower errors with fewer episodes. For example, on the task of scene-segmentation, the number of episodes required to achieve a loss of 0.006, is approximately half the number of episodes when using the proposed method than when using the IL policy. We can also see that despite IL +~\cite{mcduff2018visceral} have longer episodes on average, affect-based reaches convergence faster.

\begin{table}[t]
\centering
\begin{tabular}{lcc}  
\toprule
Method & Frame Rest. & Sketch-to-Img. \\
\midrule
Random & 276.1 & 273.6 \\ 
Straight & 260.9 & 271.4 \\
IL & 275.4 & 276.3 \\
IL +~\cite{mcduff2018visceral} & 248.7 & 260.7 \\
Affect-based & \textbf{242.7} & \textbf{257.8} \\
\toprule 
\end{tabular}
\caption{FID scores calculated for the image generation tasks. The FID was calculated on 2000 test images. We compute the metric between the reconstruction and the ground truth. The results are averaged over 30 runs.}
\label{tab:fid}
\vspace{-0.5cm}
\end{table}

Finally, besides the L2 loss, we also examine the realism of the output using the Frechet Inception Distance (FID)~\cite{heusel2017gans}, a metric that is frequently used to evaluate the realism of images generated using GANs~\cite{goodfellow2014generative}. The results are presented in Table~\ref{tab:fid} and show that better FID scores are obtained using the proposed framework.

\noindent{\bf How efficient is the framework?}
Given that our reward mechanism requires additional computation, we also consider the performance penalty we might be paying. 
We conducted timed runs for each policy and logged the average frame rate. Our code is implemented using TensorFlow 2.0, with CUDA 10.0 and CUDNN 7.6. For our framework, we observed an average of $16.3$ fps using a mobile GTX1060 and $28.9$ fps using an RTX2080Ti, which was only slightly worse than $19.1$ fps and $41.6$ fps respectively, using the IL policy. The results were averaged over $100$ seconds of driving.


\section{Discussion and Conclusions}

This paper explores how using positive affect to model an intrinsic motivation for an agent can spur exploration. Greater exploration by the agent can lead to a better representation of the environment and this, in turn, leads to improved performance in a range of downstream tasks. We argued that positive affect is an important drive that spurs 
more curious behavior. Modeling positive affect as an intrinsically motivated reward led to an exploration policy with 51\% longer duration, 46\% greater coverage and 29\% fewer collisions in comparison to IL. Fig.~\ref{fig:heatmap} shows this as a heat map. 

Central to our argument is that affective responses to stimuli are intrinsic sources of feedback that lead to exploration and discovery of examples that generalize across contexts. We used our general representations to perform experiments across multiple perception tasks. Comparing performance with and without our affective reward we found a large benefit in using the policy with intrinsic motivation based on the positive affect signal. Qualitative examples (see Fig.~\ref{fig:samples}) show that this led to a better reconstruction of the respective outputs across the tasks. Furthermore, exploration was larger using positive affect as a reward signal than when using a physiological fight-or-flight response~\cite{mcduff2018visceral} which fits with our hypothesis that positive affect is one drive for curiosity.

Here we were not attempting to mimic affective processes. But rather to show that functions trained on affect like signals can lead to improved performance. Smiles are complex nonverbal behaviors that are both common and nuanced; while smiles are interpreted as expressing positive emotion they communicate a variety of interpersonal states~\cite{ekmanFeltFalseMiserable1982,rychlowskaFunctionalSmilesTools2017}. This work establishes the potential for affect like mechanisms in robotics. Extension to other physiological signals presents a further opportunity worth exploring.


\section*{APPENDIX}
\label{sec:appendix}

\subsection{Network Architectures}
\label{ssec:networkarchitectures}

Table \ref{tab:nav_arch} shows the network architecture for the driving policy described in the paper. The architecture contains three convolutional layers and two dense layers. The input shape is $[84\times84\times4]$ (four consecutive grayscale images). The output is a vector of probabilities with a shape equal to the number of possible actions.

Table \ref{tab:afb_arch} shows the network architecture for the affect-based reward function described in the paper. The architecture contains three convolutional layers and two dense layers. Batch normalization is applied prior to each intermediate layer. The input shape is $[84\times84\times3]$ and contains a single RGB image. The output is a vector of two values. The first is the positive affect-based value that is being used by our affect-based policy, and the second is the proposed reward output from \cite{mcduff2018visceral} which is being examined using our model architecture in the fourth baseline described in the experiments.


\begin{table}[h]
\centering
\begin{tabular}{lccc}
\toprule
Layer & Act. & Out. shape & Parameters \\
\midrule
Conv2D & ReLU & $16\times8\times8$ & 4.1k \\ 
Conv2D & ReLU & $32\times4\times4$ & 8.2k \\ 
Conv2D & ReLU & $32\times3\times3$ & 9.2k \\ 
Dense & ReLU & 256 & 401k \\ 
Dense & Softmax & 5 & 1.2k \\
\midrule
Total & & & 424k \\
\toprule 
\end{tabular}
\caption{CNN architecture for the navigation policy.}
\label{tab:nav_arch}
\end{table}

\begin{table}[h]
\centering
\begin{tabular}{lccc}  
\toprule
Layer & Act. & Out. shape & Parameters \\
\midrule
Conv2D & ReLU & $32\times5\times5$ & 2.4k \\ 
Conv2D & ReLU & $48\times4\times4$ & 24.6k \\ 
Conv2D & ReLU & $64\times4\times4$ & 49.2k \\ 
Dense & ReLU & 2048 & 8.4M \\ 
Dense & Linear & 2 & 4k \\
\midrule
Total & & & 8.4M \\
\toprule 
\end{tabular}
\caption{CNN architecture for the affect-based reward function.}
\label{tab:afb_arch}
\end{table}

Table \ref{tab:vae_arch} shows the network architecture for the VAE model described in the paper. The encoder and the decoder composed of five layers each, with batch normalization prior to each intermediate layer. The input shape is $[64\times64\times3]$ and contains a single RGB image. The output of the encoder is an $8$-dimensional latent space representation. The output shape of the decoder is $[64\times64\times3]$ for a single RGB/segmentation image, or $[64\times64\times1]$ for a depth estimation map. 

\begin{table}[h]
\centering
\begin{tabular}{lccc}  
\toprule
Layer & Act. & Out. shape & Params \\
\midrule
\multicolumn{4}{l}{Encoding Layers} \\
Conv2D & ReLU & $64\times4\times4$ & 3.1k \\ 
Conv2D & ReLU & $128\times4\times4$ & 131k \\ 
Conv2D & ReLU & $256\times4\times4$ & 524k \\ 
Dense & ReLU & 1024 & 9.4M \\ 
Dense & Linear & 16 & 16.4k \\
\midrule
\multicolumn{4}{l}{Decoding Layers} \\
Dense & ReLU & 1024 & 9.2k \\ 
Dense & ReLU & 6272 & 6.4M \\ 
Conv2D Transpose & ReLU & $128\times4\times4$ & 262k \\ 
Conv2D Transpose & ReLU & $64\times4\times4$ & 131k \\ 
Conv2D Transpose & Sigmoid & $3\times4\times4$ & 3k \\ 
\midrule
Total & & & 16.9M \\
\toprule 
\end{tabular}
\caption{Architecture for the convolutional VAE model.}
\label{tab:vae_arch}
\end{table}

For archive:
For more details about the training procedure and parameters, our code is publicly available on \href{https://github.com/microsoft/affectbased}{https://github.com/microsoft/affectbased}.
For submission:
For more details about the training procedure and parameters, the code is available on the supplemental material.

\subsection{Reward Multiplication Factor}
\label{ssec:rewardmultiplicationfactor}

\begin{figure}[h]
\centering
\includegraphics[width = 0.5\textwidth]{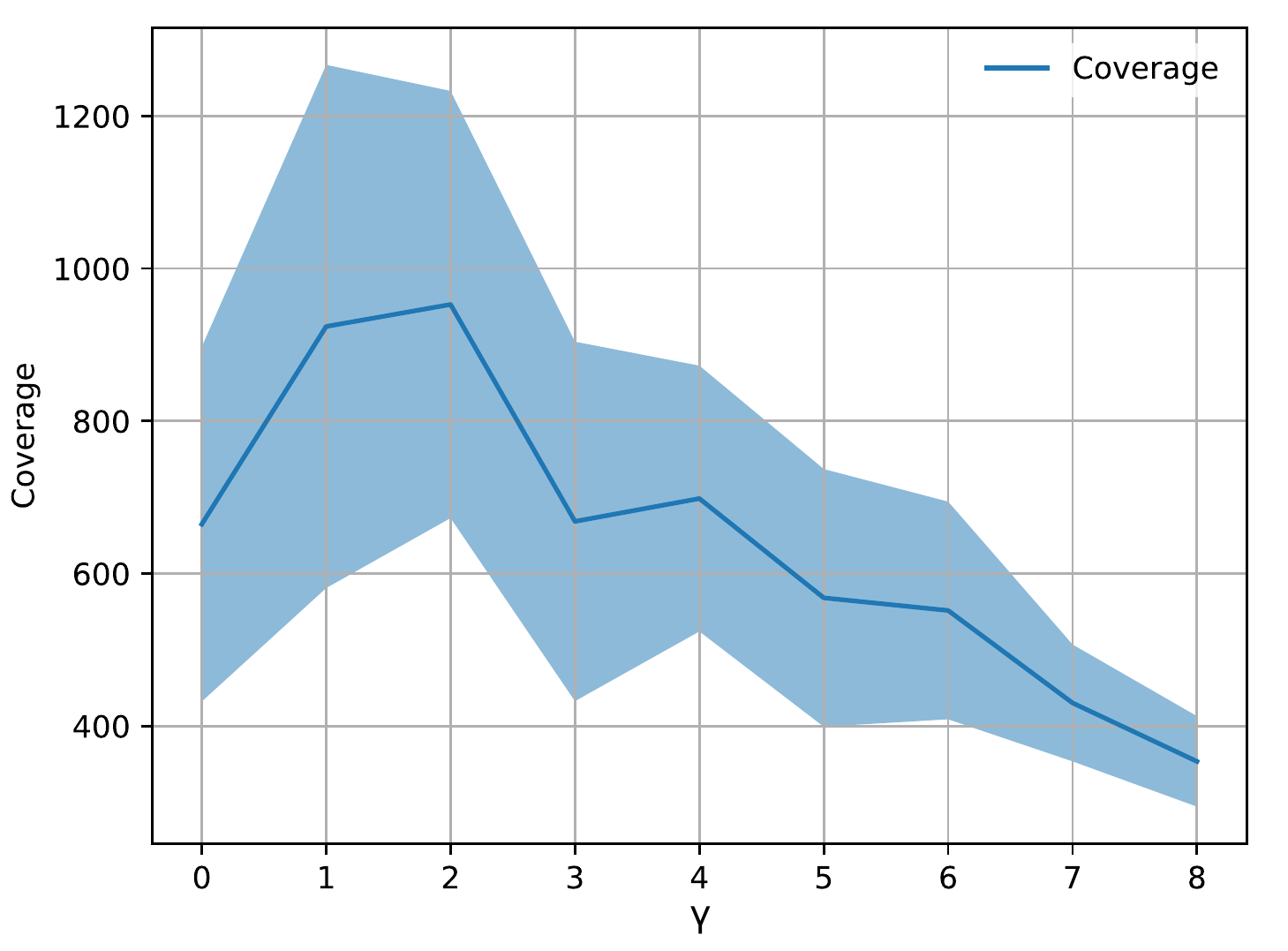}
\caption{Mean coverage as a function of gamma when searching for the right gamma for IL + \cite{mcduff2018visceral}. The plot shows that $\gamma=2$ resulted in the highest coverage.}
\label{fig:gamma_searching}
\end{figure}

Our method relies on adding the reward component on top of the action probabilities such that it will maximize the exploration results. To find the best $\gamma$, we performed the experiment shown in the coverge table that is in the paper for a range of potential values, and then we used this $\gamma$ to perform the rest of the experiments. An example of an experiment to find the $\gamma$ can be seen in Fig.~\ref{fig:gamma_searching}.

\section*{ACKNOWLEDGMENT}

We thank the subjects that helped us creating the dataset for the affect-based reward model, and the rest of the team for the help with the simulation and the experiments.


{\small
\bibliographystyle{IEEEtran}
\bibliography{main}
}

\end{document}